%% file: main.tex
\documentclass[journal]{IEEEtran}
\usepackage{amsmath, amssymb, graphicx}
\usepackage{physics2}
\usepackage{stfloats}
\usepackage{multirow}
\usepackage{tikz}
\usetikzlibrary{shapes,arrows,positioning}

\usephysicsmodule{ab}
\usepackage[inkscapelatex=false]{svg}
\usepackage{cite}
\usepackage{svg}
\usepackage{geometry}
\usepackage{titlesec}
\usepackage{hyperref}
\usepackage{setspace}
\usepackage{subcaption}

\usepackage{indentfirst}
\usepackage{multicol}

\title{Adaptive kernel-density approach for imbalanced binary classification}

\author{Kotaro~J.~Nishimura,%
        Yuichi~Sakumura,%
        and~Kazushi~Ikeda%
\thanks{K.~J.~Nishimura is with the Graduate School of Science and Technology, Nara Institute of Science and Technology, Japan. (email: nishimura.kotarojames.ni4@is.naist.jp)}%
\thanks{Y.~Sakumura is with the Data Science Center, and also with the Graduate School of Science and Technology, Nara Institute of Science and Technology, Japan. (email: saku@bs.naist.jp)\textbf{Corresponding author.}}%
\thanks{K.~Ikeda is with the Graduate School of Science and Technology and the Data Science Center, Nara Institute of Science and Technology, Japan. (email: kazushi@is.naist.jp).}%
}

\date{\today}

\begin{document}

\maketitle
\begin{abstract}
Class imbalance is a common challenge in real-world binary classification tasks, often leading to predictions biased toward the majority class and reduced recognition of the minority class. This issue is particularly critical in domains such as medical diagnosis and anomaly detection, where correct classification of minority classes is essential. Conventional methods often fail to deliver satisfactory performance when the imbalance ratio is extremely severe. To address this challenge, we propose a novel approach called Kernel-density-Oriented Threshold Adjustment with Regional Optimization (KOTARO), which extends the framework of kernel density estimation (KDE) by adaptively adjusting decision boundaries according to local sample density. In KOTARO, the bandwidth of Gaussian basis functions is dynamically tuned based on the estimated density around each sample, thereby enhancing the classifier’s ability to capture minority regions. We validated the effectiveness of KOTARO through experiments on both synthetic and real-world imbalanced datasets. The results demonstrated that KOTARO outperformed conventional methods, particularly under conditions of severe imbalance, highlighting its potential as a promising solution for a wide range of imbalanced classification problems.

\end{abstract}

\input{introduction_v2}

\input{method_v2}
\input{result_v2}
\input{discussion_v2}

\input{acknowledgment}

\bibliographystyle{IEEEtran}
\bibliography{references}
\end{document}

%% file: introduction_v2.tex
\section{Introduction}

In real-world binary classification tasks, class imbalance is a common and challenging issue, where one class significantly outnumbers the other. This phenomenon occurs in domains such as medical diagnosis, where rare diseases represent the minority class, and fraud detection, where fraudulent transactions form only a small fraction of the data. When trained on such imbalanced datasets, classifiers tend to favor the majority class, as conventional loss functions primarily optimize overall accuracy \cite{he_learning_2009,johnson_survey_2019,haixiang_learning_2017}. Consequently, minority class recognition deteriorates, and models may even achieve deceptively high accuracy by predicting only the majority class \cite{krawczyk_learning_2016,he_learning_2009}. This undermines the utility of machine learning in critical applications, where misclassification of minority samples has severe practical consequences.

To address class imbalance, preprocessing methods such as oversampling and undersampling are commonly employed. Oversampling increases the number of minority class samples, with the Synthetic Minority Over-sampling Technique (SMOTE) being a representative method that generates synthetic examples from existing ones. In contrast, undersampling reduces majority class samples, either randomly or selectively, as in NearMiss and Tomek Links, which refine decision boundaries by focusing on samples near the minority class \cite{chawla_smote_2002,batista_study_2004}.

Another line of approaches modifies the learning process itself. Class-weighted loss functions assign larger penalties to minority misclassifications, while Focal Loss further emphasizes difficult cases \cite{elkan_foundations_2001,lin_focal_2018,sun_cost-sensitive_2007}. Ensemble-based methods have also been proposed, such as Balanced Random Forest, which trains each tree on balanced subsets, and EasyEnsemble, which combines undersampling with bagging to improve minority recognition \cite{chen_using_2004,liu_exploratory_2009}. Alternative strategies involve framing the problem as anomaly detection, where methods like One-Class SVM and Isolation Forest learn to separate minority patterns from the background distribution \cite{scholkopf_estimating_2001,liu_isolation_2008}.

Despite these advances, significant limitations remain. Oversampling can introduce redundancy and distort nonlinear data structures, while undersampling risks discarding critical majority information \cite{batista_study_2004,jo_class_2004}. Class weighting requires careful parameter tuning and fails in regions where no minority samples exist \cite{bunkhumpornpat_safe-level-smote_2009,fernandez_smote_2018}. Moreover, ensemble and anomaly detection methods often incur high computational costs, limiting their scalability. Crucially, most existing approaches lack the ability to adapt decision boundaries to local density variations, even though such adaptability is key for handling extreme imbalance\cite{lopez_insight_2013}.

\begin{figure*}[t]
    \centering
    \includegraphics[width=0.95\linewidth]{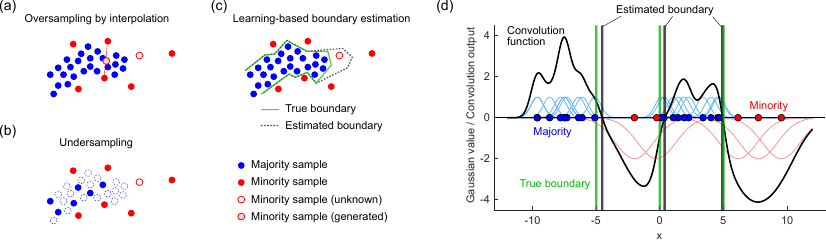}
    \caption{
Imbalanced samples and algorithms.
(a) Data structure distortion due to oversampling.
(b) Information loss due to undersampling.
(c) Limitations of boundary estimation through learning in region without samples.
(d) The proposed method uses a kernel dependent on local sample density, represented in one dimensional space for clarity.
    }
    \label{fig:concept}
\end{figure*}

In this study, we propose a novel method, Kernel-density-Oriented Threshold Adjustment with Regional Optimization (KOTARO), which extends the idea of kernel density estimation (KDE) to imbalanced classification. While conventional KDE employs a fixed Gaussian bandwidth for all samples \cite{silverman_density_nodate,terrell_variable_1992}, KOTARO dynamically adjusts the bandwidth according to local sample density. In high-density regions dominated by the majority class, the kernels shrink to tighten the boundary, whereas in sparse minority regions, the kernels expand to preserve sensitivity. Our main contributions are as follows:
\begin{itemize}
    \item We introduce a density-adaptive kernel framework for imbalanced classification.
    \item We demonstrate its effectiveness on synthetic datasets (3D–9D) under varying imbalance ratios.
    \item We validate KOTARO on real-world imbalanced datasets \cite{andre_m_carrington_new_2020}, showing superior performance over existing methods, particularly in extreme imbalance scenarios.
\end{itemize}

%% file: method_v2.tex
\section{Method}

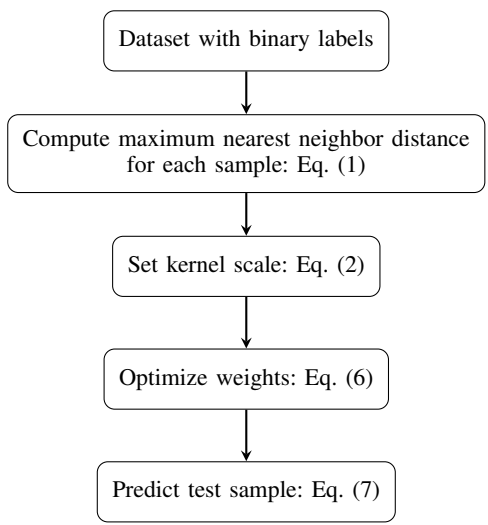
\begin{figure}[t]
\centering
\begin{tikzpicture}[node distance=1.5cm, auto]
  \tikzstyle{startstop} = [rectangle, rounded corners, draw, text centered, font=\small, minimum width=2.5cm, minimum height=0.8cm, inner sep=2mm] 
  \tikzstyle{process}   = [rectangle, rounded corners, draw, text centered, font=\small, minimum width=2.5cm, minimum height=0.8cm, inner sep=2mm]
  \tikzstyle{arrow} = [thick, ->, >=stealth]
  
  \node (start) [startstop] {Dataset with binary labels};
  \node (scale) [process, below of=start, align=center] {Compute maximum nearest neighbor distance\\
  for each sample: Eq. (\ref{eq:ditance})};
  \node (kernel) [process, below of=scale, align=center] {Set kernel scale: Eq. (\ref{eq:kernel_def})};
  \node (opt) [process, below of=kernel, align=center] {Optimize weights: Eq. (\ref{eq:opt_w})};
  \node (pred) [process, below of=opt, align=center] {Predict test sample: Eq. (\ref{eq:prediction})};
  
  \draw [arrow] (start) -- (scale);
  \draw [arrow] (scale) -- (kernel);
  \draw [arrow] (kernel) -- (opt);
  \draw [arrow] (opt) -- (pred);
\end{tikzpicture}
\caption{Workflow of the KOTARO method.}
\label{fig:workfolow}
\end{figure}

\newcommand{\bx}{\mathbf{x}}

\subsection{Core algorithm}
The KOTARO method proposed in this study is a classification algorithm inspired by kernel density estimation (KDE). 
It dynamically adjusts the kernel bandwidth (i.e., radius) according to local sample density (Fig.~\ref{fig:concept}d). 
Specifically, in high-density regions dominated by majority samples, kernels with small bandwidths are assigned positive weights, 
which emphasize the local effect of majority samples and prevent the decision boundary from expanding outward. 
Conversely, in sparse regions where minority samples are located, kernels with large bandwidths are assigned negative weights, 
so that the influence of minority samples is diffused over a broader area, enabling the classifier to recognize them more reliably. 
Moreover, by employing broader bandwidths in such sparse regions, KOTARO can partially cover areas where no samples are present, 
thereby reducing the risk of overlooking the minority class. 
The signed, density-adaptive superposition of these kernels yields a decision function with sharpened boundaries in majority regions 
and improved sensitivity to minority samples.

This mechanism can be intuitively understood by analogy with binary image processing. 
In binary images, the boundaries of pixel regions with value 1 are sharply defined. 
When a Gaussian filter is applied, however, the boundaries become blurred due to the diffusion of intensity values outward from the region of ones. 
If the filter has a small radius (i.e., narrow bandwidth), this diffusion is limited, and the boundary remains relatively sharp. 
Similarly, in KOTARO, applying narrow-bandwidth kernels in majority-dense regions suppresses the outward diffusion of their influence, 
while broader kernels in sparse regions preserve the visibility of minority samples near the boundary.

\begin{figure}[t]
    \centering
    \includegraphics[width=0.95\linewidth]{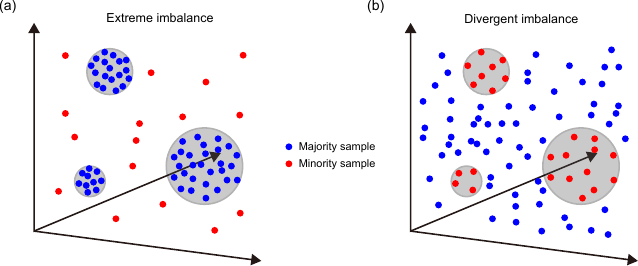}
    \caption{
Conceptual diagram of artificial imbalance data for evaluating the performance of binary classifiers.
(a) Extreme imbalance: The majority label samples are concentrated in small areas.
(b) Divergent imbalance: The overall density is not skewed, but the minority samples are limited to small areas.
}
    \label{fig:image_of_synthetic_data}
\end{figure}

The overall workflow of KOTARO can be summarized in the following steps:
\begin{enumerate}
\item For each sample point, calculate the Euclidean distances to its $n$ nearest neighbors and select the maximum distance among them.
\item Use the maximum distance obtained in step 1 as the bandwidth (i.e., radius) of the Gaussian kernel. 
Assign positive or negative weights to the kernels according to the class labels $\{-1, 1\}$, and construct the signed superposition of Gaussians.
\item Optimize the weights and threshold values using the training labels to construct the decision function $f(x)$. 
The classification boundary is then defined as the set of points satisfying $f(x)=0$.
\end{enumerate}

In step 1, the distance to the nearest neighbor is sensitive to local fluctuations, making it unstable as a scale indicator. 
Instead, we compute the distances to the $n$ nearest neighbors and adopt the maximum distance as the bandwidth. 
According to extreme value theory, the distribution of these maxima converges to a Weibull distribution, providing statistical stability. 
Consequently, even if two sample groups are located in different regions, their maximum-distance distributions are expected to be statistically similar 
when their local densities are comparable, making the maximum distance a robust measure of local density.
evaluating the similarity of local sample densities.

\subsubsection{Definition of adaptive kernel function}
Consider $N$ samples of $M$-dimensional data ${\bx_i}=[x_{i1}, \cdots,x_{iM}] \ (i=1,\cdots,N)$. 
First, define the Euclidean distance between sample ${\bx}_i$ and sample ${\bx}_j$ as 
$d(i, j)= \Vab{{\bx_i}-{\bx_j}}_2$, and let the set of $n$ nearest neighbors for the $i$-th sample be $\mathcal{N}_n(i)$. 
Then, compute
\begin{equation}
d_i = \max_{j \in \mathcal{N}_n(i)} d(i, j).
\label{eq:ditance}
\end{equation}
Using this maximum distance $d_i$, the adaptive kernel function centered at the $i$-th sample is defined as
\begin{equation}
k(\bx,\bx_i) = \exp \pab{- \gamma_i\Vab{{\bx}-{\bx_i}}^2},
\label{eq:kernel_def}
\end{equation}
where $\gamma_i=1/d_i$. 
Note that the function defined here does not necessarily guarantee the properties of Mercer kernels, 
such as symmetry $k(\bx_i,\bx_j)=k(\bx_j,\bx_i)$ or positive semidefiniteness of the kernel matrix. 
Instead, it is employed as a KDE-inspired basis function in the proposed method.

\subsubsection{Learning algorithm and label prediction for unknown samples}

The discriminant function $f(\bx)$ is defined using the adaptive kernel (basis) functions $k(\bx_i,\bx)$ and the weights $w_i$ as
\begin{equation}  
   f(\bx) = \sum^N_{i=1} w_i k(\bx_i,\bx). 
\end{equation}  

The weights $w_i$ are learned so that this discriminant function satisfies the following relation for the binary labels $y_j=\{-1,1\}$:  
\begin{equation}  
   y_j = \sum^N_{i=1} w_i k(\bx_i,\bx_j).    \label{eq:def_y}
\end{equation}  

Using the kernel matrix $K \in \mathbb{R}^{N \times N}$, where $K_{ij} = k(\mathbf{x}_i, \mathbf{x}_j)$, Eq.~(\ref{eq:def_y}) can be written as  
\begin{equation}  
   \mathbf{y} = K\mathbf{w},  \label{eq:def_y_mat}
\end{equation}  
where the label vector is $\mathbf{y}=[y_1,\cdots,y_N]^T$, and the weight vector is $\mathbf{w}=[w_1,\cdots,w_N]^T$ ($T$ denotes transpose).  
From Eq.~(\ref{eq:def_y_mat}), the weight vector $\mathbf{w}$ is obtained as  
\begin{equation}  
   \mathbf{w} = K^{-1} \mathbf{y},  
   \label{eq:opt_w}
\end{equation}  
where $K^{-1}$ denotes the inverse or, when $K$ is indefinite, the Moore--Penrose pseudoinverse. 
(Optionally, ridge regularization $K+\lambda I$ may also be applied to ensure numerical stability.)

Finally, the predicted label $y_{\text{pred}}$ for an unknown test sample $\bx_{\text{test}}$ is determined on the basis of the sign of the discriminant function:  
\begin{equation}
y_{\text{pred}} = \text{sign}(f(\bx_{\text{test}})), \quad 
\text{sign}(z) = 
\begin{cases}
 1 & (z > 0), \\
-1 & (z \leq 0).
\end{cases}
\label{eq:prediction}
\end{equation}

In this way, the KOTARO method locally adjusts the bandwidth of Gaussian basis functions, sharpening the decision boundary in majority-dense regions while preserving the sensitivity to minority samples in sparse regions. 
The overall workflow is illustrated in Fig.~\ref{fig:workfolow}.

\begin{table*}[ht]
\centering
\caption{
Composition of each real-world dataset.
}
\label{tab:exp_datasets}
\begin{tabular}{lcccc}
\hline
Dataset & \# Feature dimension  & $M_i$ & $M_a$ & $M_i/M_a$ \\ \hline
Fertility & 10 & 12 & 88 & 0.136 \\
Parkinson & 23 & 48 & 147 & 0.327 \\
Lung cancer & 56 & 9 & 23 & 0.391 \\ 
Pima & 8 & 268 & 500 & 0.536 \\\hline
\end{tabular}
\end{table*}

\subsubsection{Synthetic dataset}

To evaluate the performance of the proposed method, two types of imbalanced synthetic datasets were generated: 
(1) extreme imbalance (EI), in which the majority samples are densely concentrated in a narrow region (Fig.~\ref{fig:image_of_synthetic_data}(a)), 
and (2) divergent imbalance (DI), in which the minority samples are densely concentrated in a narrow region (Fig.~\ref{fig:image_of_synthetic_data}(b)).

The data generation procedure was as follows:
\begin{enumerate}
\item The feature space was defined as $\Omega = \prod_{i=1}^{n} [0, 5]$ with $n=2, 3, 6,$ and $9$.
\item Within this space, hyperspheres were placed. Their centers were set inside $\Omega$, and their radii were randomly determined in the range $(0, 1]$.
\item The number of hyperspheres was set to two for the two-dimensional case and three for the higher-dimensional cases.
\item Overlapping between hyperspheres and protrusion beyond $\Omega$ were allowed, but all generated samples were constrained to lie within $\Omega$.
\end{enumerate}

In the EI dataset, samples generated inside the hyperspheres were labeled as the majority class, whereas in the DI dataset they were labeled as the minority class. 
Let $M_a$ denote the number of majority samples and $M_i$ the number of minority samples. 
The imbalance ratio $M_i/M_a$ was varied within the range $(0,1]$ to construct the training datasets. 
Test samples were generated independently from the same distributions.

\subsubsection{Real-world datasets}
The real-world datasets used in this study are Fertility, Parkinson, Lung Cancer, and Pima, obtained from the UCI Machine Learning Repository \cite{noauthor_citation_nodate,little_parkinsons_2007,zq_hong_lung_1991,kahn_diabetes_0,david_gil_fertility_2012}. 
All of these datasets are imbalanced, with $M_i/M_a$ ratios ranging from $[0.136, 0.536]$ (Table~\ref{tab:exp_datasets}).

\subsubsection{Feature selection with Boruta}
To investigate the effect of feature selection, we applied the Boruta algorithm to each real-world dataset before training. 
All predictors were z-score normalized, and Boruta was executed with a Random-Forest backend (500 trees, maxIter = 100, significance level $\alpha = 0.05$). 
Only features labelled as confirmed were retained, while tentative features were discarded to keep the pipeline deterministic. 
The curated feature sets were then provided to KOTARO and all baseline classifiers. 
Performance was evaluated using stratified 5-fold cross-validation repeated ten times, and the mean G-mean, F1, and Precision scores are reported in Table \ref{tab:real_data_boruta}. 
This protocol isolates the effect of feature selection while keeping the training and evaluation schedule identical to that used for Table~\ref{tab:realdata}.

\subsubsection{Evaluation procedure}
For synthetic datasets, 50 new samples per class were generated from the true regions for evaluation. 
Training, sampling, and testing were repeated 20 times, and the mean accuracy and standard error were calculated. 
For the real-world datasets, classification performance was assessed using G-mean, F1-score, and Precision as evaluation metrics\cite{luque_impact_2019}.

G-mean evaluates the balance between Sensitivity and Specificity, which is crucial in imbalanced settings:
\[
\text{G-mean} = \sqrt{\text{Sensitivity} \times \text{Specificity}},
\]
where 
$\text{Sensitivity} = TP/(TP + FN)$ and 
$\text{Specificity} = TN/(TN + FP)$. 
F1-score balances Precision and Recall, defined as
\[
\text{F1-score} = \frac{2 \times \text{Precision} \times \text{Recall}}{\text{Precision} + \text{Recall}}, 
\]
where 
$\text{Precision} = TP/(TP + FP)$ and 
$\text{Recall} = TP/(TP + FN)$. 
Precision itself is explicitly given by
\[
\text{Precision} = \frac{TP}{TP + FP}.
\]

\subsubsection{Baseline methods}
For comparison, we evaluated the proposed method against several baselines: 
SVM with SMOTE, SVM with random undersampling, Standard SVM, Weighted SVM, One-Class Random Forest \cite{liu_exploratory_2009}, and AdaBoost. 
SVM employed a Gaussian kernel in all cases. 
The hyperparameters of each model were optimized by grid search with cross-validation, and identical training and test datasets were used for all methods.

%% file: result_v2.tex
\section{Result}


\subsection{Boundary comparison using artificial data}

\begin{figure*}[t]
  \centering
  \includegraphics[width=1.0\linewidth]{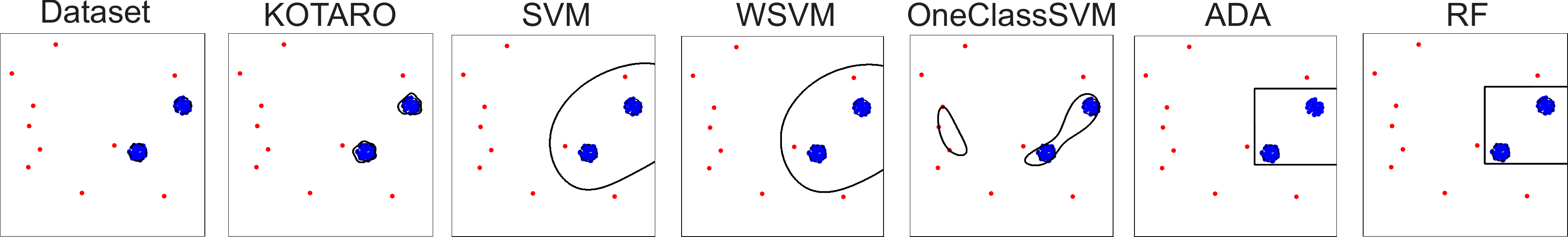}
  \caption{
Comparison of classification boundaries for the EI dataset in two dimensions. Blue points indicate the majority class, and red points indicate the minority class. The total number of samples is 100, with a majority-to-minority ratio of 9:1.
  }
  \label{fig:EI2Dboundary}

\vspace{1em} 

  \centering
  \includegraphics[width=1.0\linewidth]{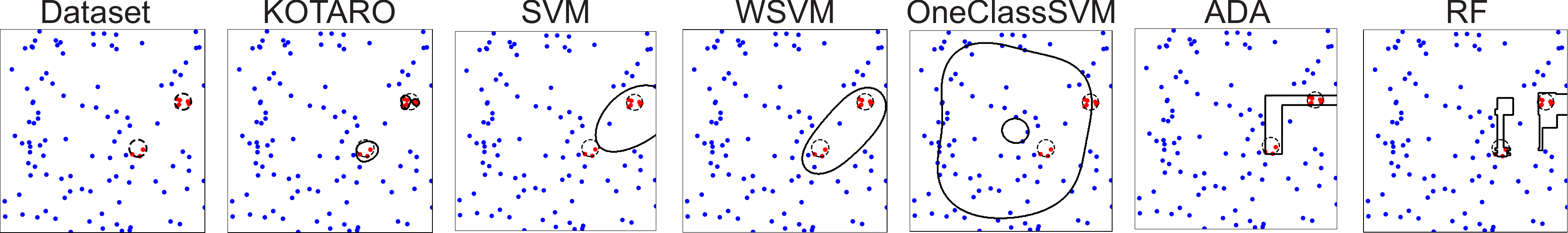}
  \caption{
Comparison of classification boundaries for the DI dataset in two dimensions.
  }
  \label{fig:DI2Dboundary}
\end{figure*}

To illustrate the effectiveness of the proposed method, we first compared the decision boundaries obtained on a two-dimensional dataset with an extreme imbalance (EI) (Fig.~\ref{fig:image_of_synthetic_data}a), where the majority samples are densely concentrated in narrow regions. 
Figure~\ref{fig:EI2Dboundary} visualizes the decision boundaries learned by different classifiers. 
The proposed method successfully delineated the minority class while preserving the compact shapes of the majority regions. 
In contrast, most baseline methods tended to over-extend the majority class boundaries, thereby encroaching on areas that should be assigned to the minority class. 
Although One-Class SVM produced relatively tight boundaries, it mistakenly merged two separate majority regions and even included regions with no majority samples. 
These observations highlight that the proposed method can construct sharper and more faithful decision boundaries for EI-type imbalanced data, capturing the true structure of the underlying distributions more effectively than conventional approaches.

Next, decision boundaries were compared for the DI-type imbalanced dataset, where minority class samples are confined to narrow regions (Fig.~\ref{fig:DI2Dboundary}). 
The proposed method produced boundaries that were sometimes slightly wider or narrower than the true ones but overall captured the correct structure and yielded satisfactory separations. 
In contrast, most baseline methods incorrectly merged the two minority regions into a single area, resulting in poor accuracy. 
Random Forest managed to separate the two regions but substantially overestimated their sizes. 
One-Class SVM, which is inherently biased toward learning the blue positive class, produced the most inappropriate boundaries. 
These results demonstrate that the proposed method can also construct suitable and reliable decision boundaries for DI-type imbalanced data.

\begin{figure*}[t]
  \centering
  \includegraphics[width=0.9\linewidth]{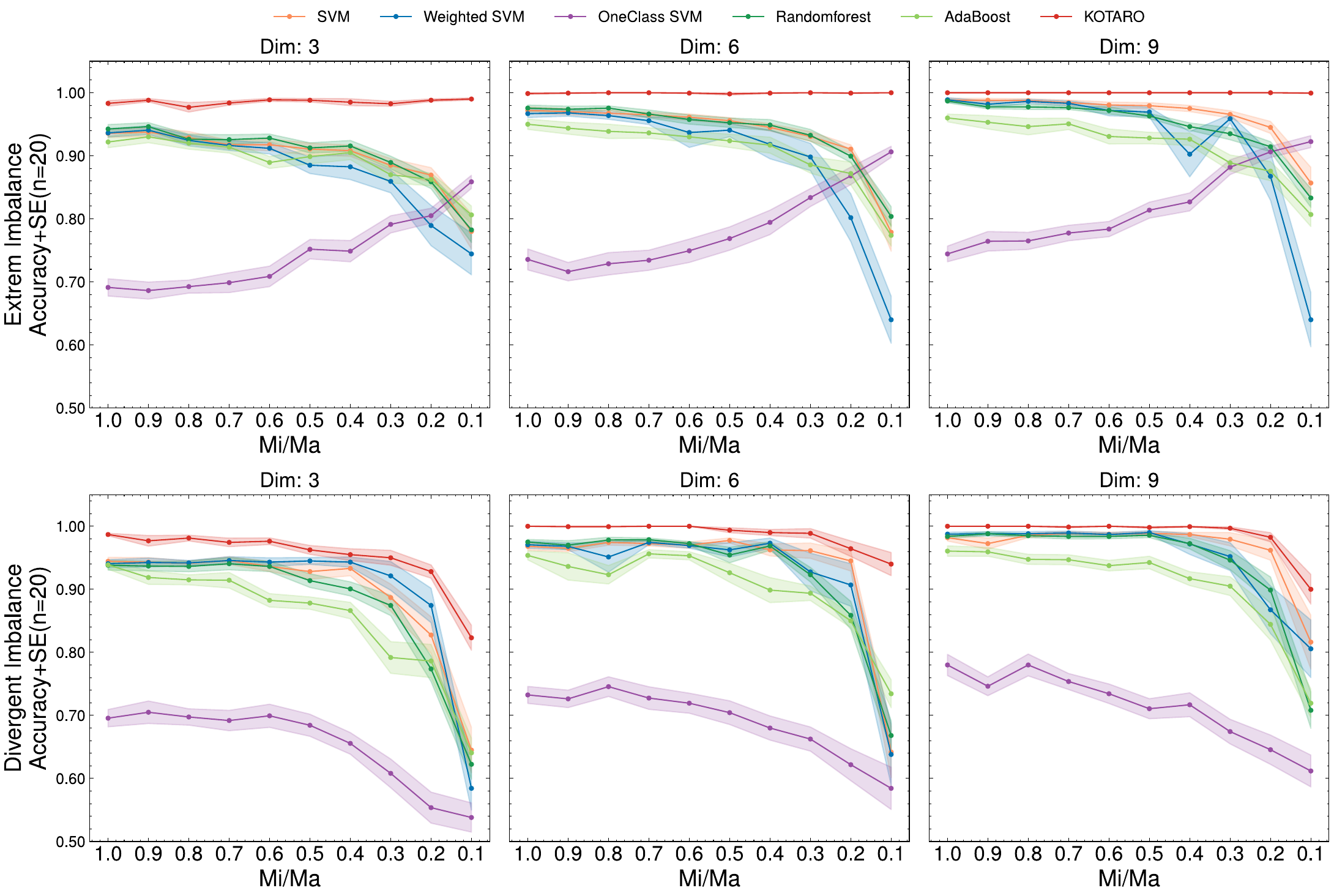}
  \caption{
Binary classification accuracy for imbalanced high-dimensional synthetic datasets. 
The upper and lower rows correspond to EI-type and DI-type imbalance, respectively, and from left to right the graphs show results for three-, six-, and nine-dimensional samples. 
The vertical axis denotes classification accuracy, while the horizontal axis indicates the imbalance ratio ($M_i / M_a$), where smaller values correspond to stronger imbalance. 
Because the test set was balanced (50 positive + 50 negative samples), always predicting the positive class yields 50\% accuracy as a baseline. 
Each curve shows the mean accuracy with standard errors computed over 20 independent trials.
}
  \label{fig:accuracy_comparison}
\end{figure*}

\begin{table*}[b]
\centering
\caption{
Performance comparison using real datasets in Table \ref{tab:exp_datasets}. RUS = Random Undersampling; WSVM = Weighted SVM; RF = Random Forest; ADA = AdaBoost. The bold values indicate the best performance for each metric among the algorithms.
}
\label{tab:realdata}
\begin{tabular}{ccccccccc}
\hline  & & KOTARO  & SMOTE+SVM  & RUS+SVM  & SVM  & WSVM & RF & ADA \\ \hline
\multirow{2}{*}{Fertility} & G-mean  & $\bf{0.621}$ & 0.391 & 0.379 & 0 & 0 & 0  & 0  \\
  & F1-score & 0.200 & $\bf{0.224}$ & 0.156 & 0 & 0 & 0  & 0  \\
\hline
\multirow{2}{*}{Parkinson} & G-mean  & 0.656 & 0.522 & 0.392 & 0.278 & 0.552 & 0.810 & $\bf{0.811}$ \\
  & F1-score & 0.507 & 0.443 & 0.285 & 0.151 & 0.342 & $\bf{0.772}$ & 0.753 \\
\hline
\multirow{2}{*}{Lung Cancer} & G-mean  & 0.397 & $\bf{0.461}$ & 0.311 & 0 & 0 & 0  & 0  \\
  & F1-score & $\bf{0.440}$ & 0.417 & 0.281 & 0 & 0.440 & 0  & 0  \\
\hline
\multirow{2}{*}{Pima} & G-mean  & 0.591 & $\bf{0.722}$ & 0.719 & 0.681 & 0.719 & 0.723 & 0.680 \\
  & F1-score & 0.501 & $\bf{0.645}$ & 0.644 & 0.603 & 0.642 & 0.650 & 0.600 \\
 \hline
\end{tabular}
\end{table*}
\begin{table*}[b]
\centering
\caption{Performance comparison on real datasets with Boruta Feature Selection 
RUS = Random Undersampling; WSVM = Weighted SVM; RF = Random Forest; ADA = AdaBoost.
Boldface highlights the best score for each metric across all algorithms.}
\label{tab:real_data_boruta}
\begin{tabular}{ccccccccc}
\hline
                             &          & KOTARO       & SMOTE+SVM & RUS+SVM      & SVM   & WSVM         & RF           & ADA   \\ \hline
\multirow{2}{*}{Fertility}   & G-mean   & $\bf{0.745}$ & 0.370     & 0.543        & 0     & 0            & 0.279        & 0     \\
                             & F1-score & $\bf{0.300}$   & 0.167     & 0.216        & 0     & 0            & 0.233        & 0     \\ \hline
\multirow{2}{*}{Parkinson}   & G-mean   & 0.115        & 0.876     & 0.741        & 0.783 & 0.836        & $\bf{0.900}$ & 0.880 \\
                             & F1-score & 0.813        & 0.913    & 0.884        & 0.917 & 0.925        & $\bf{0.956}$ & 0.942 \\ \hline
\multirow{2}{*}{Lung cancer} & G-mean   & 0.774        & 0.720     & $\bf{0.844}$ & 0     & 0.742        & 0.711        & 0.683 \\
                             & F1-score & 0.667        & 0.627     & $\bf{0.760}$ & 0     & 0.655        & 0.621        & 0.602 \\ \hline
\multirow{2}{*}{Pima}        & G-mean   & 0.703        & 0.735     & $\bf{0.737}$ & 0.725 & 0.732        & 0.713        & 0.700 \\
                             & F1-score & 0.632        & 0.660     & 0.663        & 0.660 & $\bf{0.689}$ & 0.636        & 0.625 \\ \hline
\end{tabular}
\end{table*}

\subsection{Accuracy comparison using artificial data}

Based on the result that the proposed method estimated relatively good decision boundaries for two-dimensional imbalanced samples, we evaluated its classification performance on high-dimensional synthetic data (Fig.~\ref{fig:accuracy_comparison}). 
Specifically, the proposed method was compared with five existing classification models on imbalanced samples with three, six, and nine dimensions. 
Except for One-Class SVM, all models showed a tendency for classification accuracy to decrease as the imbalance became stronger, regardless of whether the dataset was of the EI or DI type. 
In contrast, One-Class SVM exhibited the unique characteristic that its classification accuracy improved as the imbalance increased, but only for the EI-type dataset.

For the EI-type dataset (Fig.~\ref{fig:accuracy_comparison}), the proposed method, SVM, and Random Forest showed nearly identical performance on three-dimensional samples. 
However, as the dimensionality increased, the superiority of the proposed method became more pronounced. 
Notably, under highly imbalanced conditions ($M_i / M_a = 0.1$), whereas the classification accuracies of SVM and Random Forest dropped significantly, the proposed method maintained high accuracy. 
In particular, Random Forest showed the most substantial decrease in accuracy under this condition.

\begin{figure*}[t]
  \centering
  \includegraphics[width=0.9\linewidth]{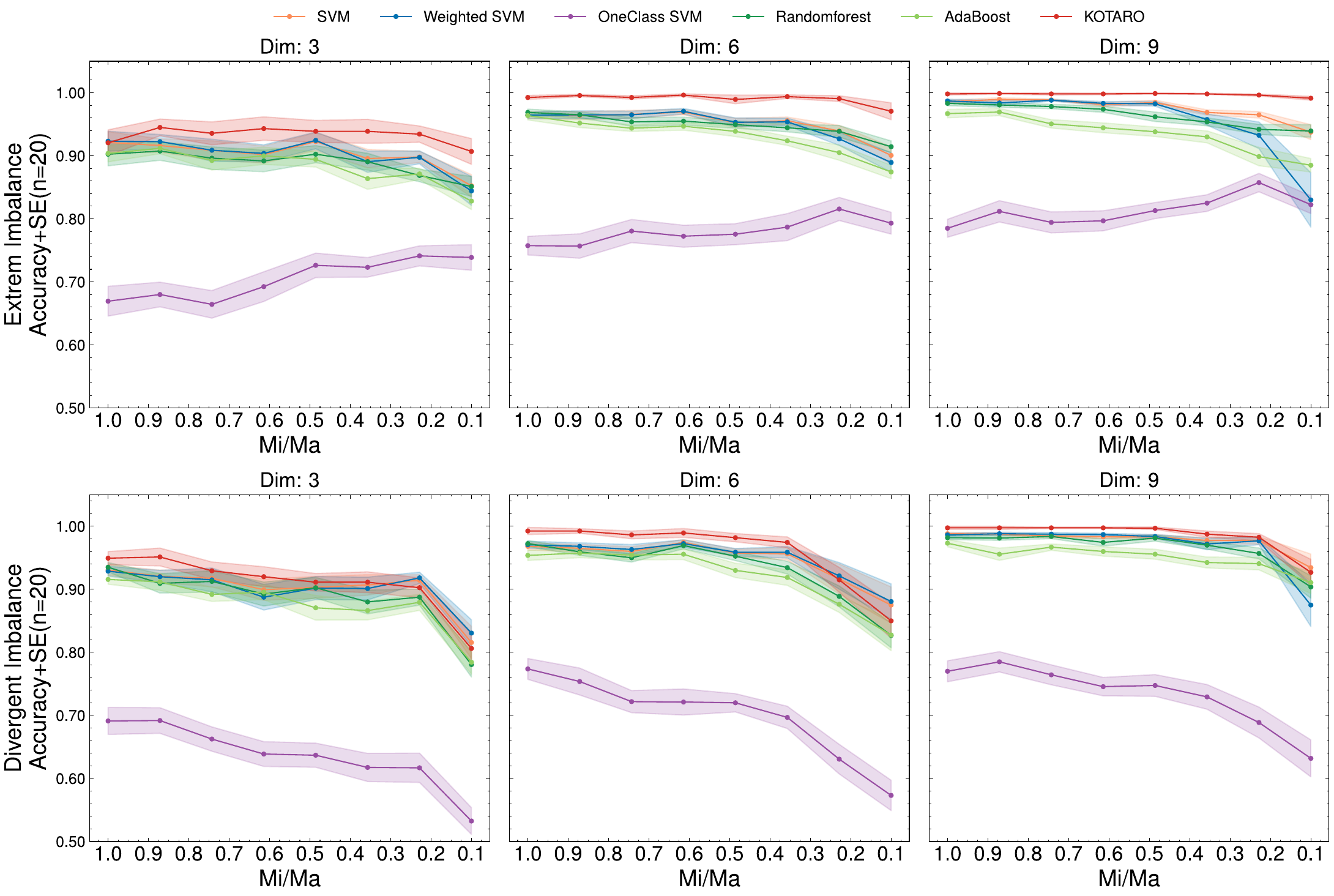}
  \caption{
Accuracy comparison using datasets where the dimensionality was doubled by adding noise dimensions, followed by feature selection with Boruta. 
The title of each figure indicates the original number of dimensions. 
The values of the noise dimensions were sampled from uniform random numbers independent of the class boundaries.
}
  \label{fig:accuracy_comparison_with_noise}
\end{figure*}

In the DI-type dataset (Fig.~\ref{fig:accuracy_comparison}), the proposed method, SVM, and Random Forest also exhibited similar performance. 
Even in the six- and nine-dimensional cases, SVM achieved performance comparable to that of the proposed method. 
These results demonstrate that the proposed method consistently achieves high classification accuracy regardless of the type of imbalanced dataset.

In general, not all dimensions contribute to classification. 
Therefore, it is necessary to evaluate classification performance when noise dimensions are added, as in the setting of Fig.~\ref{fig:accuracy_comparison_with_noise}. 
In this study, independent noise dimensions were constructed by sampling features from a uniform random distribution unrelated to the class regions and appending them to the original dimensions. 
This procedure doubled the total dimensionality, producing datasets with six, twelve, and eighteen dimensions. 
Feature selection was then applied using Boruta, and only the dimensions estimated to be important were retained. 
The selected dimensions did not necessarily correspond to the original ones \cite{kursa_feature_2010}. 
The classification performance of all algorithms was then evaluated on these selected-dimension datasets (Fig.~\ref{fig:accuracy_comparison_with_noise}). 
Under these conditions, the KOTARO method outperformed the others particularly for EI-type datasets with strong class imbalance. 
For DI-type datasets as well, it rarely performed worse than the other methods.

\subsection{Accuracy comparison using real-world data}
Finally, classification performance characteristics were compared using imbalanced real-world datasets related to diseases (Table \ref{tab:exp_datasets}). 
Unlike in Figs.~\ref{fig:EI2Dboundary} and \ref{fig:DI2Dboundary}, where balanced test samples can be generated, real-world data do not allow for such control. 
Therefore, five-fold cross-validation was used for prediction. 
G-mean and F1-score were employed as evaluation metrics (Table \ref{tab:realdata}). 
The results showed that the proposed method achieved superior classification accuracy on the Fertility dataset, which had the highest imbalance ratio. 
Similarly, a relatively high classification accuracy was obtained for the Parkinson and Lung Cancer datasets, both of which also exhibited high imbalance ratios. 
In contrast, on the Pima dataset, which had a lower imbalance ratio, the classification accuracy decreased compared with the other datasets.

\subsection{Accuracy comparison using real-world data with feature selection}
After introducing Boruta feature selection (Table \ref{tab:real_data_boruta}), the performance hierarchy shifts markedly. 
KOTARO retains clear supremacy only on the most severely imbalanced Fertility dataset. 
On Parkinson, Random Forest rises to first place, while RUS + SVM outperforms all competitors on both Lung Cancer and Pima. 
These trends suggest that once noisy dimensions are pruned, ensemble or re-sampled margin-based learners can exploit the cleaned feature space more effectively, whereas KOTARO’s density-adaptive kernel remains advantageous mainly under conditions of extreme imbalance and residual noise.

%% file: discussion_v2.tex
\section{Discussion}

\subsection*{Effectiveness of the proposed method KOTARO}
In this study, we proposed KOTARO, a method that constructs decision boundaries on the basis of the local density of samples labeled with two classes, and demonstrated that it can improve classification accuracy under severe class imbalance (Fig.~\ref{fig:EI2Dboundary}). 
The primary reason for the decline in classification accuracy on imbalanced data is that decision boundaries are strongly influenced by majority class samples, resulting in an overestimation of the majority class regions. 
The proposed method leverages the characteristic that majority class samples are densely distributed, and by reducing the kernel size in high-density regions, it adaptively adjusts the boundary to prevent the majority class regions from “bleeding out.”

\subsection*{Analogy based on image processing}
This behavior can be explained using an analogy from image processing. 
When a Gaussian filter is applied to a binary image to produce a blurred image, a smaller Gaussian size results in weaker blurring. 
Moreover, adaptive thresholding methods that depend on brightness are sometimes used when binarizing blurred images. 
The proposed method can be interpreted as effectively performing threshold adjustment by varying the degree of blurring.

\subsection*{Superiority on EI-type imbalanced data}
The reason why the proposed method outperformed the other methods on EI-type imbalanced data lies in its handling of regions without samples. 
In EI-type minority class regions, large portions often contain very few or no samples. 
SVM and Random Forest do not recognize these regions as minority class areas; instead, they construct decision boundaries aimed at minimizing overall misclassification. 
As a result, the minority class regions become narrower and tend to be underestimated.

In contrast, the proposed method constructs decision boundaries based on regions where majority class samples are present. 
Consequently, it can recognize unsampled areas as minority class regions. 
This property allows the proposed method to appropriately estimate minority class regions in EI-type imbalanced data, thereby improving classification accuracy.

\subsection*{Analysis of performance on DI-type imbalanced data}
On the other hand, no significant difference in classification accuracy was observed between the proposed method and the other methods on balanced and DI-type imbalanced datasets. 
In particular, for DI-type imbalanced datasets, the performance difference between the proposed method and SVM was negligible (Figs.~\ref{fig:accuracy_comparison}, \ref{fig:accuracy_comparison_with_noise}). 
This is because the density difference between majority and minority class samples is small. 
In such cases, the proposed method effectively employs a uniform kernel size, making its decision boundary nearly identical to that of SVM with a Gaussian kernel. 
Therefore, it would be effective to estimate in advance whether the sample distribution exhibits uneven density and to apply the proposed method when the majority class exhibits high density.

\subsection*{Applicability to real-world data}
Real-world datasets, such as those in the medical field, are often imbalanced, making the proposed method highly applicable. 
Indeed, the proposed method demonstrated high classification accuracy on medical datasets with severe imbalance used in this study (see Table \ref{tab:exp_datasets}). 
In contrast, for the Pima dataset, the accuracy was lower, probably because of its weak imbalance and the possibility that the features themselves do not contain sufficient discriminative information. 
Since other classification models also achieved only about 60\% accuracy on this dataset, accuracy improvements could be expected by incorporating additional features or employing feature selection methods, such as the Wrapper method.

%
%
\subsection*{Impact of feature selection on real-world performance}
The baseline comparison in Table \ref{tab:realdata} confirms the earlier narrative: KOTARO dominates when imbalance is both severe and skew-dense. 
It leads on Fertility and remains highly competitive on Parkinson and Lung-Cancer, while all methods plateau on the mildly imbalanced Pima set.

Once Boruta feature selection is introduced (Table \ref{tab:real_data_boruta}), the ranking reshuffles. 
KOTARO still prevails on Fertility, but ensemble-style and re-sampled margin learners (Random Forest on Parkinson; RUS + SVM on Lung-Cancer and Pima) overtake after noisy dimensions are removed. 
This shift indicates that the density-adaptive kernel retains its edge chiefly when (i) class imbalance is extreme, (ii) majority samples cluster tightly, and (iii) residual feature noise persists despite selection. 
In more moderate settings with cleaner feature spaces, classical models close the gap or surpass KOTARO.

Overall, Tables \ref{tab:realdata}--\ref{tab:real_data_boruta} together suggest a two-phase strategy: deploy KOTARO as the default for raw, highly imbalanced data, then reassess after feature curation—switching to ensembles or re-sampled SVMs if they show superior post-selection generalization.

\subsection*{Future directions}
Two main directions are suggested for future research. 
First, accuracy validation under conditions closer to real-world scenarios is needed. 
Real-world data distributions are often skewed, resembling log-normal distributions. 
Therefore, evaluating performance using synthetic data sampled from log-normal distributions instead of uniform distributions would enhance the practical applicability of the proposed method. 
Second, developing an automatic algorithm for distinguishing between EI-type and DI-type imbalanced data is important. 
Since the proposed method showed high classification accuracy for EI-type imbalanced data, an algorithm capable of automatically identifying whether a given imbalanced dataset is EI- or DI-type would expand the applicability of the proposed method and further enhance its practical value.

%% file: acknowledgment.tex
\section*{Acknowledgments}
We would like to express our gratitude to K.T. Suzuki for  valuable discussions and insightful advice that greatly contributed to this study.
This work was supported in part by Nara Institute of Science and Technology. Data Science Center, Grants for new issue identification activities. 
\section*{Author contributions}
Conceptualization: Y.S. and K.I.; Methodology: K.J.N. and Y.S.; Investigation: K.J.N.; Data Analysis: K.J.N.; Writing – Original Draft: K.J.N. and Y.S.; Writing – Review \& Editing: K.I.; Visualization: K.J.N.; Supervision: K.I.; Funding Acquisition: K.I.; Resources: K.I.

%% file: main.bbl
\begin{thebibliography}{10}
\providecommand{\url}[1]{#1}
\csname url@samestyle\endcsname
\providecommand{\newblock}{\relax}
\providecommand{\bibinfo}[2]{#2}
\providecommand{\BIBentrySTDinterwordspacing}{\spaceskip=0pt\relax}
\providecommand{\BIBentryALTinterwordstretchfactor}{4}
\providecommand{\BIBentryALTinterwordspacing}{\spaceskip=\fontdimen2\font plus
\BIBentryALTinterwordstretchfactor\fontdimen3\font minus \fontdimen4\font\relax}
\providecommand{\BIBforeignlanguage}[2]{{%
\expandafter\ifx\csname l@#1\endcsname\relax
\typeout{** WARNING: IEEEtran.bst: No hyphenation pattern has been}%
\typeout{** loaded for the language `#1'. Using the pattern for}%
\typeout{** the default language instead.}%
\else
\language=\csname l@#1\endcsname
\fi
#2}}
\providecommand{\BIBdecl}{\relax}
\BIBdecl

\bibitem{he_learning_2009}
\BIBentryALTinterwordspacing
H.~He and E.~A. Garcia, ``Learning from {Imbalanced} {Data},'' \emph{IEEE Transactions on Knowledge and Data Engineering}, vol.~21, no.~9, pp. 1263--1284, Sep. 2009. [Online]. Available: \url{https://ieeexplore.ieee.org/document/5128907/}
\BIBentrySTDinterwordspacing

\bibitem{johnson_survey_2019}
\BIBentryALTinterwordspacing
J.~M. Johnson and T.~M. Khoshgoftaar, ``Survey on deep learning with class imbalance,'' \emph{Journal of Big Data}, vol.~6, no.~1, p.~27, Mar. 2019. [Online]. Available: \url{https://doi.org/10.1186/s40537-019-0192-5}
\BIBentrySTDinterwordspacing

\bibitem{haixiang_learning_2017}
\BIBentryALTinterwordspacing
G.~Haixiang, L.~Yijing, J.~Shang, G.~Mingyun, H.~Yuanyue, and G.~Bing, ``Learning from class-imbalanced data: {Review} of methods and applications,'' \emph{Expert Systems with Applications}, vol.~73, pp. 220--239, May 2017. [Online]. Available: \url{https://www.sciencedirect.com/science/article/pii/S0957417416307175}
\BIBentrySTDinterwordspacing

\bibitem{krawczyk_learning_2016}
\BIBentryALTinterwordspacing
B.~Krawczyk, ``\BIBforeignlanguage{en}{Learning from imbalanced data: open challenges and future directions},'' \emph{\BIBforeignlanguage{en}{Progress in Artificial Intelligence}}, vol.~5, no.~4, pp. 221--232, Nov. 2016. [Online]. Available: \url{https://doi.org/10.1007/s13748-016-0094-0}
\BIBentrySTDinterwordspacing

\bibitem{chawla_smote_2002}
\BIBentryALTinterwordspacing
N.~V. Chawla, K.~W. Bowyer, L.~O. Hall, and W.~P. Kegelmeyer, ``{SMOTE}: {Synthetic} {Minority} {Over}-sampling {Technique},'' \emph{Journal of Artificial Intelligence Research}, vol.~16, pp. 321--357, Jun. 2002, arXiv:1106.1813 [cs]. [Online]. Available: \url{http://arxiv.org/abs/1106.1813}
\BIBentrySTDinterwordspacing

\bibitem{batista_study_2004}
\BIBentryALTinterwordspacing
G.~E. A. P.~A. Batista, R.~C. Prati, and M.~C. Monard, ``A study of the behavior of several methods for balancing machine learning training data,'' \emph{SIGKDD Explor. Newsl.}, vol.~6, no.~1, pp. 20--29, 2004. [Online]. Available: \url{https://dl.acm.org/doi/10.1145/1007730.1007735}
\BIBentrySTDinterwordspacing

\bibitem{elkan_foundations_2001}
C.~Elkan, ``\BIBforeignlanguage{en}{The {Foundations} of {Cost}-{Sensitive} {Learning}}.''\hskip 1em plus 0.5em minus 0.4em\relax International Joint Conference on Artificial Intelligence, 2001, pp. 973--978.

\bibitem{lin_focal_2018}
\BIBentryALTinterwordspacing
T.-Y. Lin, P.~Goyal, R.~Girshick, K.~He, and P.~Dollár, ``Focal {Loss} for {Dense} {Object} {Detection},'' Feb. 2018, arXiv:1708.02002 [cs]. [Online]. Available: \url{http://arxiv.org/abs/1708.02002}
\BIBentrySTDinterwordspacing

\bibitem{sun_cost-sensitive_2007}
\BIBentryALTinterwordspacing
Y.~Sun, M.~S. Kamel, A.~K.~C. Wong, and Y.~Wang, ``Cost-sensitive boosting for classification of imbalanced data,'' \emph{Pattern Recognition}, vol.~40, no.~12, pp. 3358--3378, Dec. 2007. [Online]. Available: \url{https://www.sciencedirect.com/science/article/pii/S0031320307001835}
\BIBentrySTDinterwordspacing

\bibitem{chen_using_2004}
C.~Chen, A.~Liaw, and L.~Breiman, ``\BIBforeignlanguage{en}{Using {Random} {Forest} to {Learn} {Imbalanced} {Data}},'' Department of Statistics, University of California, Berkeley, Tech. Rep. Technical Report 666, 2004.

\bibitem{liu_exploratory_2009}
\BIBentryALTinterwordspacing
X.-Y. Liu, J.~Wu, and Z.-H. Zhou, ``Exploratory {Undersampling} for {Class}-{Imbalance} {Learning},'' \emph{IEEE Transactions on Systems, Man, and Cybernetics, Part B (Cybernetics)}, vol.~39, no.~2, pp. 539--550, Apr. 2009. [Online]. Available: \url{https://ieeexplore.ieee.org/document/4717268}
\BIBentrySTDinterwordspacing

\bibitem{scholkopf_estimating_2001}
\BIBentryALTinterwordspacing
B.~Schölkopf, J.~C. Platt, J.~Shawe-Taylor, A.~J. Smola, and R.~C. Williamson, ``Estimating the {Support} of a {High}-{Dimensional} {Distribution},'' \emph{Neural Computation}, vol.~13, no.~7, pp. 1443--1471, Jul. 2001. [Online]. Available: \url{https://ieeexplore.ieee.org/document/6790022}
\BIBentrySTDinterwordspacing

\bibitem{liu_isolation_2008}
\BIBentryALTinterwordspacing
F.~T. Liu, K.~M. Ting, and Z.-H. Zhou, ``Isolation {Forest},'' in \emph{2008 {Eighth} {IEEE} {International} {Conference} on {Data} {Mining}}, Feb. 2008, pp. 413--422, iSSN: 2374-8486. [Online]. Available: \url{https://ieeexplore.ieee.org/document/4781136}
\BIBentrySTDinterwordspacing

\bibitem{jo_class_2004}
\BIBentryALTinterwordspacing
T.~Jo and N.~Japkowicz, ``Class imbalances versus small disjuncts,'' \emph{SIGKDD Explor. Newsl.}, vol.~6, no.~1, pp. 40--49, 2004. [Online]. Available: \url{https://dl.acm.org/doi/10.1145/1007730.1007737}
\BIBentrySTDinterwordspacing

\bibitem{bunkhumpornpat_safe-level-smote_2009}
C.~Bunkhumpornpat, K.~Sinapiromsaran, and C.~Lursinsap, ``\BIBforeignlanguage{en}{Safe-{Level}-{SMOTE}: {Safe}-{Level}-{Synthetic} {Minority} {Over}-{Sampling} {TEchnique} for {Handling} the {Class} {Imbalanced} {Problem}},'' in \emph{\BIBforeignlanguage{en}{Advances in {Knowledge} {Discovery} and {Data} {Mining}}}, T.~Theeramunkong, B.~Kijsirikul, N.~Cercone, and T.-B. Ho, Eds.\hskip 1em plus 0.5em minus 0.4em\relax Berlin, Heidelberg: Springer, 2009, pp. 475--482.

\bibitem{fernandez_smote_2018}
\BIBentryALTinterwordspacing
A.~Fernandez, S.~Garcia, F.~Herrera, and N.~V. Chawla, ``\BIBforeignlanguage{en}{{SMOTE} for {Learning} from {Imbalanced} {Data}: {Progress} and {Challenges}, {Marking} the 15-year {Anniversary}},'' \emph{\BIBforeignlanguage{en}{Journal of Artificial Intelligence Research}}, vol.~61, pp. 863--905, Apr. 2018. [Online]. Available: \url{https://jair.org/index.php/jair/article/view/11192}
\BIBentrySTDinterwordspacing

\bibitem{lopez_insight_2013}
\BIBentryALTinterwordspacing
V.~López, A.~Fernández, S.~García, V.~Palade, and F.~Herrera, ``An insight into classification with imbalanced data: {Empirical} results and current trends on using data intrinsic characteristics,'' \emph{Information Sciences}, vol. 250, pp. 113--141, Nov. 2013. [Online]. Available: \url{https://www.sciencedirect.com/science/article/pii/S0020025513005124}
\BIBentrySTDinterwordspacing

\bibitem{silverman_density_nodate}
\BIBentryALTinterwordspacing
B.~Silverman, \emph{Density {Estimation} for {Statistics} and {Data} {Analysis}}, ser. Monographs on {Statistics} and {Applied} {Probability}.\hskip 1em plus 0.5em minus 0.4em\relax London: Chapman \& Hall, vol.~26. [Online]. Available: \url{https://doi.org/10.1007/978-1-4899-3324-9}
\BIBentrySTDinterwordspacing

\bibitem{terrell_variable_1992}
G.~R. Terrell and D.~W. Scott, ``Variable {Kernel} {Density} {Estimation},'' \emph{The Annals of Statistics}, vol.~20, no.~3, pp. 1236--1265, 1992.

\bibitem{andre_m_carrington_new_2020}
\BIBentryALTinterwordspacing
{André M. Carrington}, {Paul Fieguth}, {Hammad Qazi}, {Andreas Holzinger}, {Alice P. Chen}, {Franz Mayr}, and {Douglas G. Manuel}, ``\BIBforeignlanguage{en}{A new concordant partial {AUC} and partial c statistic for imbalanced data in the evaluation of machine learning algorithms},'' \emph{\BIBforeignlanguage{en}{BMC Medical Informatics and Decision Making}}, vol.~20, no.~1, Jan. 2020. [Online]. Available: \url{https://doi.org/10.1186/s12911-019-1014-6}
\BIBentrySTDinterwordspacing

\bibitem{noauthor_citation_nodate}
\BIBentryALTinterwordspacing
``Citation - {UCI} {Machine} {Learning} {Repository}.'' [Online]. Available: \url{https://archive.ics.uci.edu/citation}
\BIBentrySTDinterwordspacing

\bibitem{little_parkinsons_2007}
\BIBentryALTinterwordspacing
M.~Little, ``Parkinsons,'' 2007. [Online]. Available: \url{https://archive.ics.uci.edu/dataset/174}
\BIBentrySTDinterwordspacing

\bibitem{zq_hong_lung_1991}
\BIBentryALTinterwordspacing
J.~Y. Z.Q.~Hong, ``Lung {Cancer},'' 1991. [Online]. Available: \url{https://archive.ics.uci.edu/dataset/62}
\BIBentrySTDinterwordspacing

\bibitem{kahn_diabetes_0}
\BIBentryALTinterwordspacing
M.~Kahn, ``Diabetes,'' 0. [Online]. Available: \url{https://archive.ics.uci.edu/dataset/34}
\BIBentrySTDinterwordspacing

\bibitem{david_gil_fertility_2012}
\BIBentryALTinterwordspacing
J.~G. David~Gil, ``Fertility,'' 2012. [Online]. Available: \url{https://archive.ics.uci.edu/dataset/244}
\BIBentrySTDinterwordspacing

\bibitem{luque_impact_2019}
\BIBentryALTinterwordspacing
A.~Luque, A.~Carrasco, A.~Martín, and A.~de~las Heras, ``The impact of class imbalance in classification performance metrics based on the binary confusion matrix,'' \emph{Pattern Recognition}, vol.~91, pp. 216--231, Jul. 2019. [Online]. Available: \url{https://www.sciencedirect.com/science/article/pii/S0031320319300950}
\BIBentrySTDinterwordspacing

\bibitem{kursa_feature_2010}
\BIBentryALTinterwordspacing
M.~B. Kursa and W.~R. Rudnicki, ``\BIBforeignlanguage{en}{Feature {Selection} with the {Boruta} {Package}},'' \emph{\BIBforeignlanguage{en}{Journal of Statistical Software}}, vol.~36, pp. 1--13, Sep. 2010. [Online]. Available: \url{https://doi.org/10.18637/jss.v036.i11}
\BIBentrySTDinterwordspacing

\end{thebibliography}
